# Toward an Automated, Proactive Safety Warning System Development for Truck Mounted Attenuators in Mobile Work Zones

**Xiang Yu, P.E., PTOE**
Traffic and Transportation Engineer
Cook, Flatt & Strobel Engineers, P.A., Kansas City, MO, USA, 64131
Email: xytm4@mail.missouri.edu

**Linlin Zhang**
PhD Student
Department of Civil and Environment Engineering
University of Missouri-Columbia, Columbia, MO, USA, 65201
Email: lz5f2@mail.missouri.edu

**Yaw, Adu-Gyamfi, Ph.D.**
Associate Professor
Department of Civil and Environment Engineering
University of Missouri-Columbia, Columbia, MO, USA, 65201
Email: adugyamfiy@missouri.edu

Word Count: 4431 words + 2 table (250 words per table) = 4,931 words

*Submitted [Aug 1, 2024]*

*Yu, Zhang, and Adu-Gyamfi*

**ABSTRACT**

Even though Truck Mounted Attenuators (TMA)/Autonomous Truck Mounted Attenuators (ATMA) and traffic control devices are increasingly used in mobile work zones to enhance safety, work zone collisions remain a significant safety concern in the United States. In Missouri, there were 63 TMA-related crashes in 2023, a 27% increase compared to 2022. Currently, all the signs in the mobile work zones are passive safety measures, relying on drivers' recognition and attention. Some distracted drivers may ignore these signs and warnings, raising safety concerns. In this study, we proposed an additional proactive warning system that could be applied to the TMA/ATMA to improve overall safety. A feasible solution has been demonstrated by integrating a Panoptic Driving Perception algorithm into the Robot Operating System (ROS) and applying it to the TMA/ATMA systems. This enables us to alert vehicles on a collision course with the TMA. Our experimental setup, currently conducted in a laboratory environment with two ROS robots and a desktop GPU, demonstrates the system's capability to calculate real-time distance and speed and activate warning signals. Leveraging ROS's distributed computing capabilities allows for flexible system deployment and cost reduction. In future field tests, by combining the stopping sight distance (SSD) standards from the AASHTO Green Book, the system enables real-time monitoring of oncoming vehicles and provides additional proactive warnings to enhance the safety of mobile work zones.

## INTRODUCTION

Truck Mounted Attenuators (TMA) traffic control devices have been increasingly utilized in mobile works in recent years. These TMA devices are attached to the rear of the follower truck. Road workers perform maintenance works such as stripping and patching in the leader truck, while the follower trucks play a protective role. In a collision, the follower truck can absorb the impact of collisions and reduce its severity. Despite these protective measures, mobile work zone collisions remain a significant safety concern in the United States. According to the statistical data from the Missouri Department of Transportation (MoDOT), there were 63 TMA-related crashes in 2023, representing a 27% increase compared to 2022. MoDOT identified the primary causes of these work zone crashes as driver inattention and excessive speed.

As the application of emerging techniques in the industry, both workers and motorists can be further protected in mobile work zones. Building on the TMA, Autonomous Truck-Mounted Attenuators (ATMA) systems have been developed. These systems utilize Vehicle-to-Vehicle (V2V) communications to remove the need for a human driver in the follower truck, particularly in hazardous environments, thereby further reducing the risk of injury. Since the follower truck operates without a driver, it is equipped with various sensors to ensure it can effectively follow the leader truck and detect obstacles. These sensors normally include radar, cameras, LiDAR, and the Global Positioning System (GPS). In a sense, the follower truck is essentially an autonomous vehicle.

Currently, autonomous vehicles and various forms of robots are attracting attention from researchers and engineers. To ensure these machines can operate safely, numerous sensors such as LiDAR and cameras are installed to provide comprehensive perception of the real-world. However, having just the hardware is not enough. The Robot Operating System (ROS) is also essential, as it integrates these sensors and software, enabling them to function as a whole system. Additionally, there have been significant advancements in object detection and semantic segmentation algorithms, such as You Only Look Once (YOLO) (1), Segment Anything Model (SAM) (2), and You Only Look Once for Panoptic Driving Perception (YOLOP) (3). Among them, YOLOP can perform traffic object detection, drivable area segmentation, and lane detection. These capabilities are crucial for improving perception abilities, especially in transportation. Embedding these algorithms into the ROS will make the full utilization of various sensors, improving the system's intelligibility and performance.

In this paper, we propose an autonomous TMA or ATMA alarm system with ROS and YOLOP integrated. This system aims to enhance the intelligence and safety of work zone operations by providing comprehensive perception and autonomous decision-making capabilities. By leveraging the strengths of ROS and YOLOP, the autonomous TMA alarm system will better protect workers and motorists in dynamic and hazardous environments. Currently, the experiment is simulated in a laboratory setting, utilizing two ROS robots and a desktop GPU. Since both robots are equipped with Jetson Nano, which has limited processing capability, the desktop GPU serves as a computation center. This network was built largely benefiting from one key feature of ROS: distributed computing. One robot, with a color and depth camera installed, functions as the follower truck, while the other robot simulates as an oncoming vehicle. The desktop GPU subscribes to raw color and depth image topics from the follower truck, processes these images using the YOLOP algorithm, calculates real-time distance and speed, and publishes warning information to the follower truck as needed. The follower truck subscribes to the warning topic, and upon receiving the warning information, it activates its LED light to warn the oncoming vehicle. The whole process is displayed in **Figure 1**. In summary, the main contributions of this study are as follows:

1). Integrating YOLOP with ROS, achieving real-time processing and warning system, and employing distributed computing improves the flexibility of hardware design for TMA/ATMA in future filed tests.

2). The system's effectiveness is demonstrated in a laboratory environment, enabling real-time processing. A warning system is implemented that allows the follower truck to activate LED lights to alert oncoming vehicles based on their real-time distance and speed.

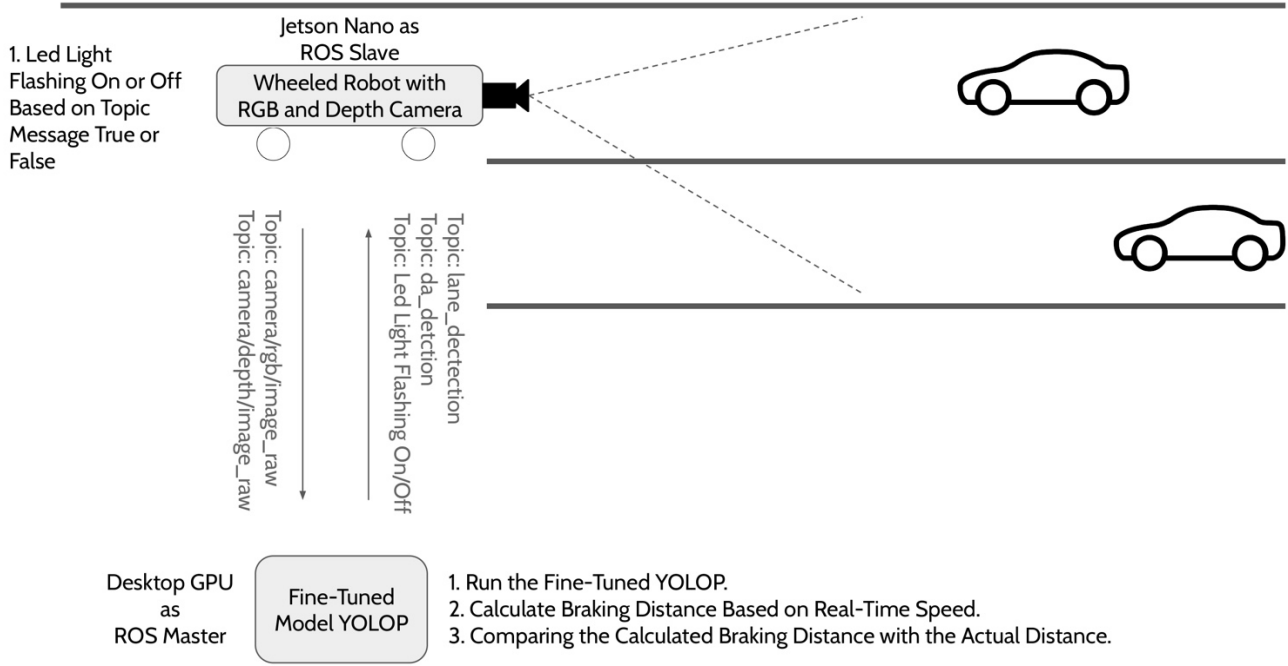

**Figure 1 Experiment Architecture in this Study**

## RELATED WORKS

In this section, we primarily focus on two domains: the current research on TMA and ATMA systems to enhance the safety of mobile work zones, and the application of the ROS in transportation work zones.

(4) indicated that the speed difference between mobile work zones and actual vehicle travel speed, as well as the failure to recognize and respond in time, are the main reasons for crashes. The most common type of crash is the rear-end (5). In Missouri alone, since 2013, there have been 80 crashes involving mobile work zones, resulting in many injuries to Department of Transportation (DOT) workers (4). To improve the safety of work zones, in recent years, TMA devices have been widely used, providing protection not only for DOT workers but also for the driver of the follower trucks. Moreover, (6) pointed out that TMA devices are highly effective in reducing the severity and cost of rear-end crashes. Even though the application of TMA reduces economic loss and provides a safer work zone environment for DOT employees, in 2023, there were 63 TMA-related crashes in Missouri, representing an increase compared to 2022. Therefore, additional safety measures are needed to further reduce TMA-related crashes. The most important of these measures is to increase driver recognition of TMAs. (7) pointed out that properly placing retroreflective chevron markings can increase TMA recognition. They also recommended positioning the amber and white flashing warning LEDs at least 1500 ft from the work zone to enhance visibility and safety. Nonetheless, incorporating traffic signages into TMA operations is considered a form of passive safety.

To enhance the safety of TMA truck drivers and leverage emerging technologies, several state Departments of Transportation (DOT), including those in Colorado, Florida, Tennessee, Missouri, and Virginia, have tested the ATMA system (5). Through field testing, which included 26 scenarios in Florida (8), 24 case scenario tests in Tennessee (9), and 31 field tests in Missouri (10), researchers identified several issues with the ATMA system. These issues included loss of connection, failure to operate within small radius curves and roundabouts, drifting off the intended path, slow movement due to GPS signal loss, and a lack of practical driving guidance. Some of these issues were addressed and resolved in their research. However, these studies focused on the system's safety for the public and workers, without exploring proactive warning systems.

(11) conducted a field test for the alarm device and directional audio system (DAS) in mobile work zones with TMA. The results indicated an increase in the distance at which vehicles begin merging and a decrease in merging speed. The alarm sound level was tested to ensure compliance with national noise standards. However, in the field test, the distance between trailing vehicles and the TMA truck was virtually estimated by the truck driver using the number of skips on the lane striping as a reference. Although the use of proactive warning systems in TMA has been explored, this approach is prone to human error, which would reduce the system's overall safety. Some studies have explored the application of proactive warning in work zones. (12) set up Light Detection and Ranging (LiDAR) sensors along roadside within work zone. These sensors perform object detection and tracking, providing predictive warning to workers. This system architecture was developed with Robot Operation System (ROS). However, this setup is within the stationary work zones.

In this study, we aim to integrate YOLOP into ROS and eventually install it in TMAs/ATMAs. This integration will provide an additional fully automated proactive safety warning system, enabling inattention drivers to become aware of the TMA/ATMA presence and take emergency actions to avoid risks.

**METHODOLOGY AND EXPERIMENT DESIGN**

We would like to achieve a fully automated active safety warning system to improve the safety of mobile work zones. The use of ROS will simplify the transition to an automated system. Additionally, the distributed computing capability will enhance the system's feasibility. Since mobile work zones typically occupy one or a few lanes, monitoring all approaching vehicles is unnecessary. Therefore, in addition to vehicle detection, additional roadway information is essential to accurately locate oncoming vehicles. YOLOP is employed because it simultaneously provides vehicle detection, driving area segmentation, and lane line segmentation. This information is crucial for determining which lanes the oncoming vehicles are in.

**Robot Operating System (ROS)**

ROS is an open-source meta-operating system for robots. It provides essential services expected of an operation system, including hardware abstraction, inter-process message passing, package management, implementation of commonly used functionalities, and low-level device control. In ROS, the most important concepts are nodes, topics, services, and parameters. Based on our study and needs, we mainly used nodes and topics. All the data processing, computation, and control are within nodes. For example, in our study, the fine-tuned YOLOP algorithm, vehicle localization, distance and speed calculations, and warning requirements are all integrated into nodes. Topics provide a publish/subscribe mechanism for nodes to communicate with each other. Nodes can publish messages to a topic or subscribe to a topic to receive messages. In our case, all raw color image data, raw depth image data, and warning information are contained in topics. The primary reason we adopted ROS in our study is its distributed computing capability. This allows different nodes to run on multiple devices, working collaboratively to complete a single task. This feature enables the flexibility of hardware design.

**You Only Look Once for Panoptic Driving Perception (YOLOP)**

YOLOP is an advanced computer vision algorithm with significant applications in transportation. It can simultaneously perform traffic object detection, drivable area segmentation, and lane line segmentation, making it a versatile tool for enhancing road safety and efficiency in autonomous driving and traffic management systems. YOLOP utilizes the YOLOv4 (13) backbone structure, CSP-Darknet (14). Compared to Darknet-53 (1), which is used in YOLOv3 (1), CSP-Darknet replaces the Leaky ReLU activation function with Mish (15) and applies CSPNet (16) to the residual blocks. In the neck of the network, Spatial Pyramid Pooling (SPP) (17) is employed to ensure the input image size is not limited and to capture more features. Additionally, the Feature Pyramid Network (FPN) (18) is applied to enhance semantic information capture and improve object detection performance.

In the object detection head, the Path Aggregation Network (PANet) (19) is used, leveraging the bottom-up pathway to improve the detection of small objects. For the driving area segmentation and lane line segmentation heads, three upsampling operations are applied to the output of the FPN in the neck. This approach ensures that the segmentation results are accurate and detailed. The YOLOP algorithm architecture is illustrated in **Figure 2**.

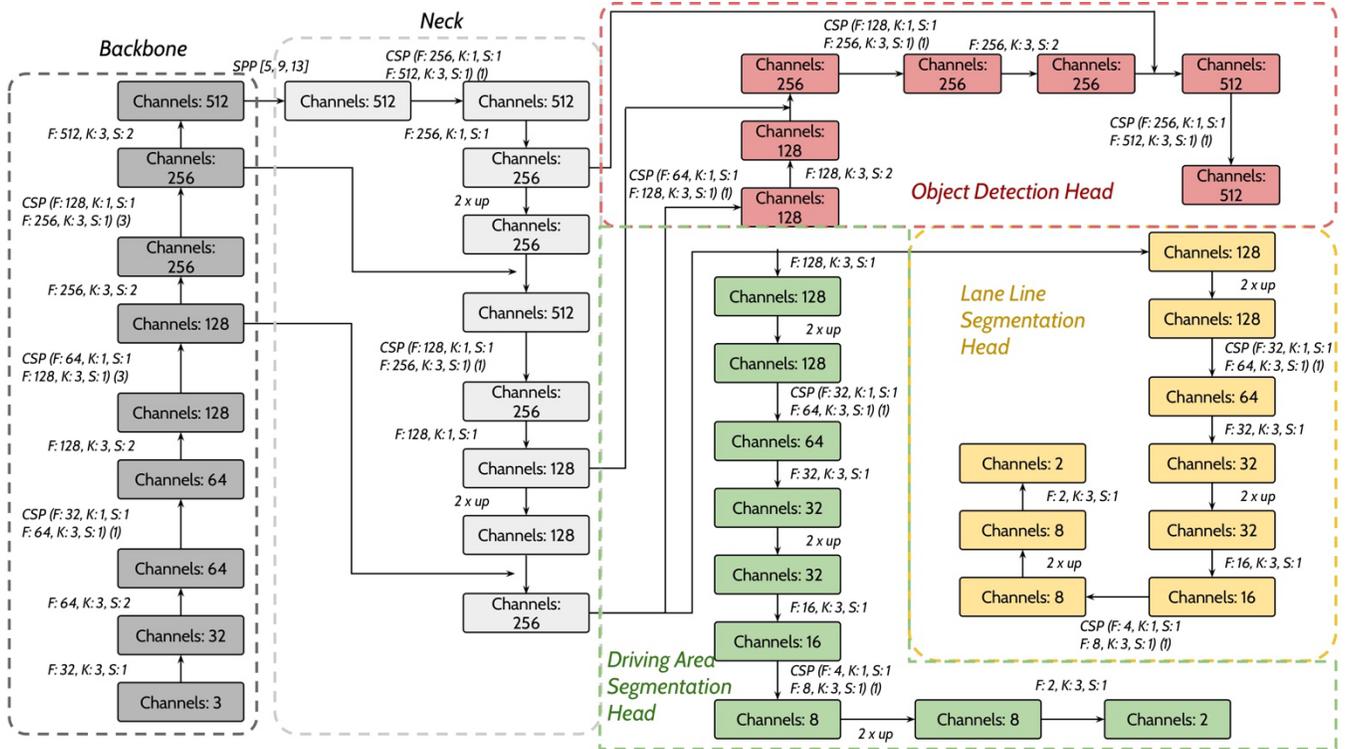

**Figure 2 You Only Look Once for Panoptic Driving Perception (YOLOP) Architecture**

In our study, we need to re-train the model since our exploration is only in the laboratory simulation environment currently. As shown in **Figure 3**, the rightmost four images represent the format of our training dataset. This includes raw color images, driving area segmentation images, lane line segmentation images, and annotated objects in JSON format. There are 415 training datasets and 100 validation datasets. There are a total of 240 training epochs. **Figure 4** shows the training total loss, driving area intersection over union (IOU), lane line IOU, and object detection average precision (AP).

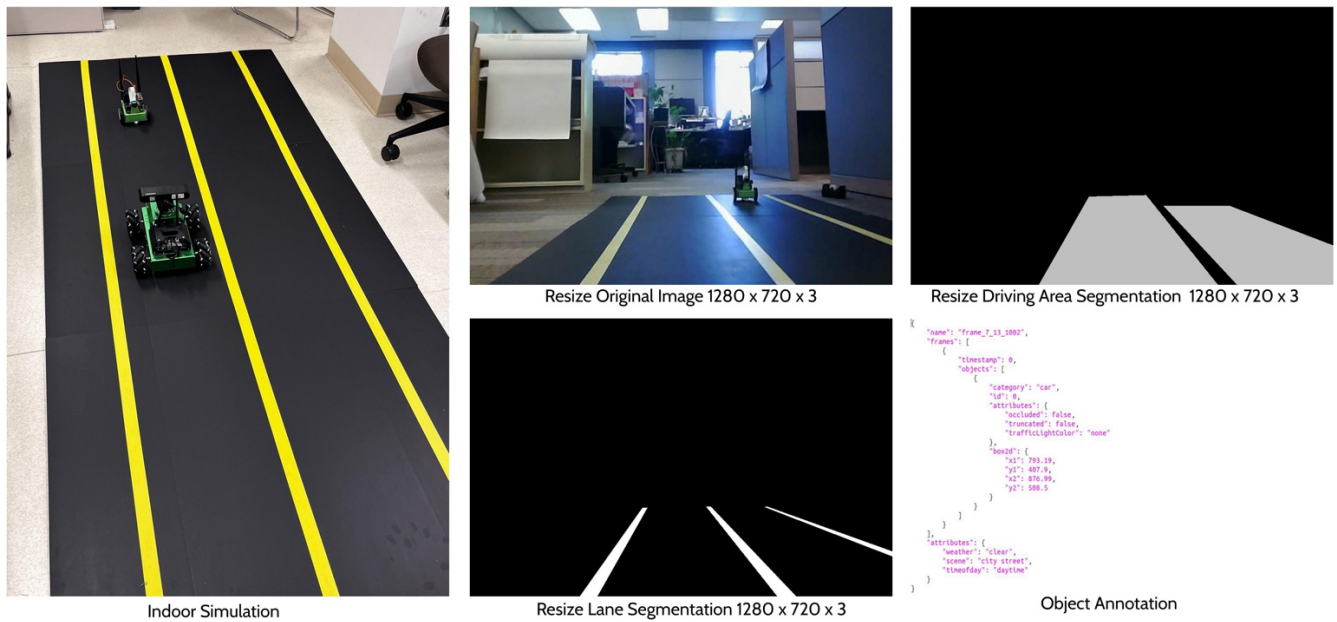

**Figure 3 Simulation Setup (Left One) and Training Dataset (Right Four)**

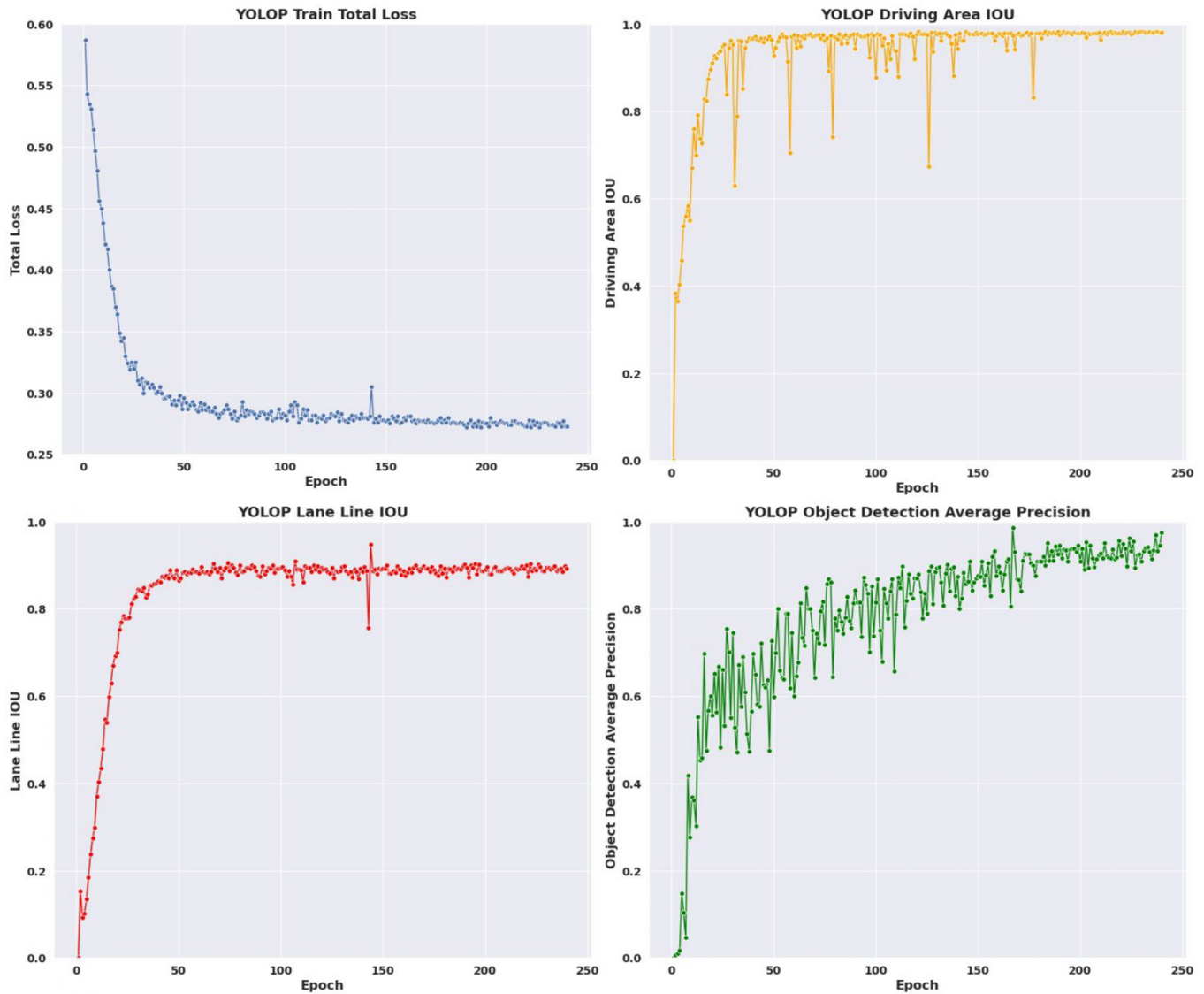

**Figure 4 Training Total Loss, Driving Area and Lane Line IOU, and Object Detection Average Precision**

**Laboratory Simulation**

In a laboratory setting, we designed a simulation involving two Jetson Nano robots. One robot, equipped with both a color camera and a depth camera, serves as the follower truck in mobile work zone with TMA attached. The other robot simulates an oncoming vehicle approaching the mobile work zone. There are two scenarios in our simulation, 1) both robots travel forward in the same lane but at different speeds; 2) the robots travel forward in the different lane, again at different speeds. Throughout the simulation, we can capture the real-time distance between the oncoming vehicle and the follower truck and further calculate the real-time speed. Additionally, the current Speed Stop Distance (SSD) can be calculated based on the real-time speed. By comparing the calculated SSD and real-time distance, the system will determine if the real-time distance is greater than the calculated SSD. If it is not, additional flashing LED lights will be activated. The simulation setting is illustrated in the leftmost image of **Figure 3**.

**System Structure**

Due to the limited processing capability of the budget-friendly Jetson Nano, directly running the fine-tuned YOLOP model on this device does not meet the real-time requirements. To address this issue, we leverage distributed computing, a key feature of ROS. In our setup, a desktop GPU serves as the ROS Master, while the Jetson Nano acts as ROS Slave. Both devices are connected to the same Wi-Fi network, allowing for efficient data transfer and processing distribution.

During operation, the follower truck, equipped with a Jetson Nano, monitors the oncoming vehicle, which is also equipped with a Jetson Nano, and publishes the camera data in real-time under the topics "/camera/rgb/image_raw" and "/camera/depth/image_raw" to the ROS network. The desktop GPU subscribes to those topics and processes the color images and the depth images using the fine-tuned YOLOP model. The model outputs object detection results, lane line segmentation results, and driving area segmentation results. Utilizing the object detection results and the data from the "/camera/depth/image_raw" topic, the desktop GPU calculates the real-time distance between the follower truck and the oncoming vehicle, and subsequently calculates the real-time speed of the oncoming vehicle. The desktop GPU also publishes warning information under the topic "/led/status" to the network, and the follower truck subscribes to this topic. The data in this topic is simple: True or False, indicating whether the warning is active or off. In the simulation, if the distance between the oncoming vehicle and follower truck is less than 0.3 meter, the data in "/led/status" topic will turn True, thereby activating the flashing warning LEDs. It is worth noting that all processing on the desktop GPU is performed within a single node.

According to the Stopping Sight Distance (SSD) guidelines (shown as **Table 1**) in the AASHTO "A Policy on Geometric Design of Highways and Streets" (Green Book), if the real-time distance is less than the SSD corresponding to the current calculated speed, the message in topic "/led/status" will turn True then the additional flashing LED warning light will be activated to alert drivers to change lanes or initiate an emergency brake. The SSD calculation equation is shown as **Equation 1**:

$$SSD = 1.47Vt + 1.075\frac{V^2}{a} \qquad (1)$$

In the equation, $SSD$ indicates stopping sight distance in $ft$, $V$ indicates the speed in $mph$, $t$ means brake reaction time, 2.5 seconds, and $a$ indicates deceleration rate in $ft/s^2$. For deceleration rate, $11.2\ ft/s^2$ is normally used.

**Table 1 Stopping Sight Distance (SSD) on Level Roadways from AASHTO "A Policy on Geometric Design of Highways and Streets" (Green Book)**

| Design Speed (mph) | Designed SSD (ft) | Design Speed (mph) | Designed SSD (ft) |
|---|---|---|---|
| 15 | 80 | 50 | 425 |
| 20 | 115 | 55 | 495 |
| 25 | 155 | 60 | 570 |
| 30 | 200 | 65 | 645 |
| 35 | 250 | 70 | 730 |
| 40 | 305 | 75 | 820 |
| 45 | 360 | 80 | 910 |

*Vehicle Localization*

After fine-tuned YOLOP algorithm outputs segmentation results, an important step is to determine which lane the oncoming vehicle is in. If the vehicle is in the same lane as the follower truck, it is necessary to continuously acquire distance and speed information. However, if the vehicle is in a different lane, to conserved computational resources, there is no need to obtain its distance or calculate its speed. The output lane line segmentation image is displayed as the top left image in **Figure 5**.

The vehicle localization code is executed immediately following the fine-tuned YOLOP outputs code. In the vehicle localization code, computer vision techniques such as median blur and connected components are implemented. The median blur is an effective image denoising technique, while the connected components technique aids in image segmentation tasks. In our study, the area threshold has been set to 500 for the connected components.

The next step is to obtain the x value corresponding to the maximum y value and the x value corresponding to the minimum y value for each connected component in the image. This allows for the calculation of the slope and intercept of each lane. As illustrated in **Figure 5**, the top right and bottom two images display these results.

The center point (x, y) of the object detection bounding box can be calculated based on the fine-tuned YOLOP outputs. By substituting the y value of the bounding box center point into the lane line equation, the corresponding x value of the lane line can be obtained. Finally, by comparing the x value of the bounding box center point with the calculated x value of the lane line, the lane in which the bounding box is located can be determined. If x_left < x_center < x_medium, then the vehicle is in left lane. If x_medium < x_center < x_right, then the vehicle is in the right lane. The results are shown in **Figure 6**.

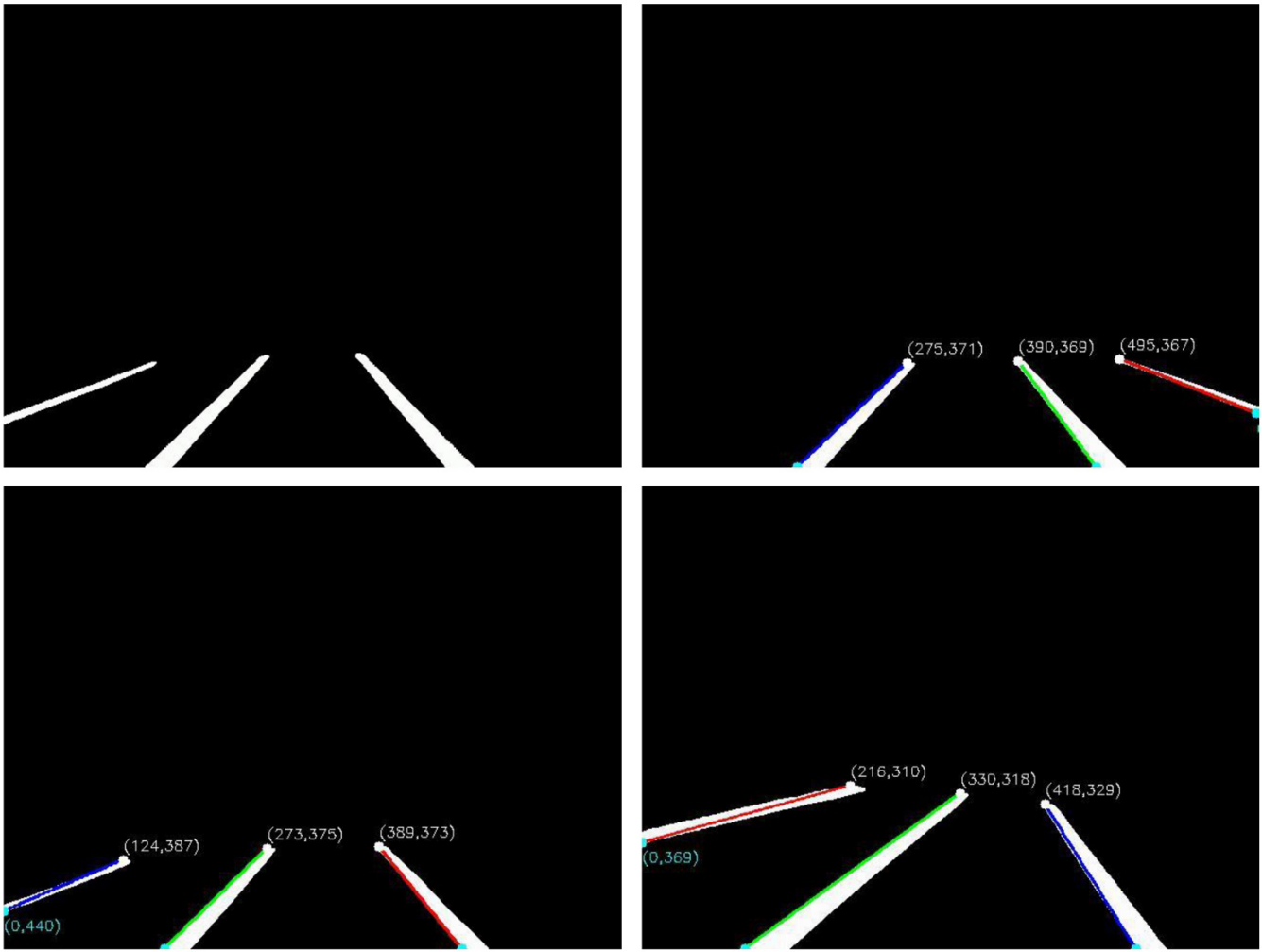

**Figure 5 Lane Function Based on the Lane Segmentation**

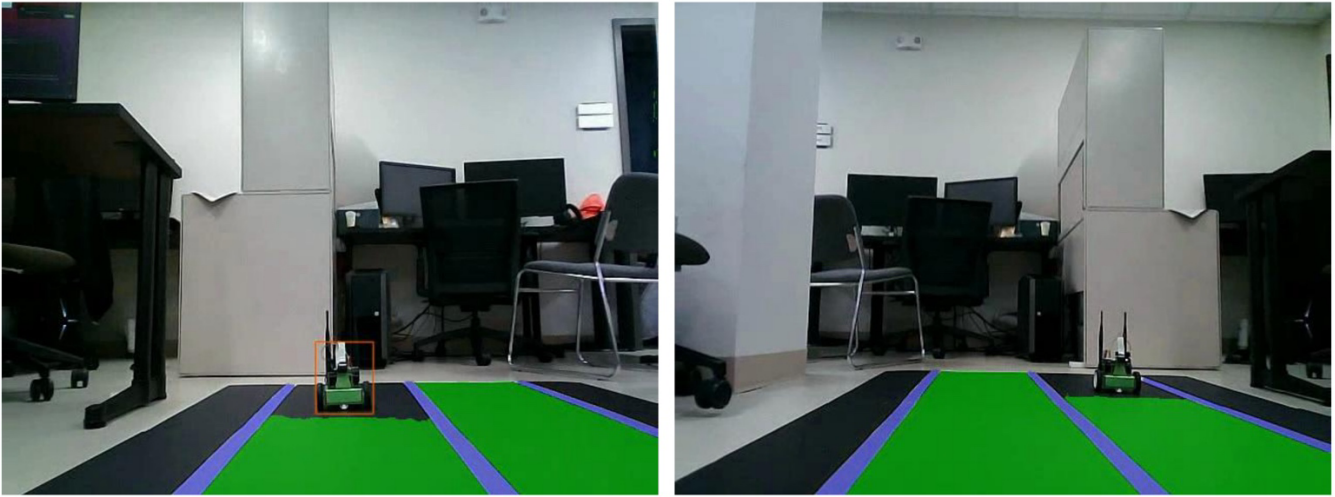

**Figure 6 Results of Vehicle Localization**

*Real-Time Distance and Speed*

The real-time distance between the oncoming vehicle and the follower truck can be calculated using the data (as shown as **Figure 7**) from the depth camera with the topic "/camera/depth/image_raw" and the object detection bounding box. To minimize the influence of the background environment on distance measurements, we calculate the average depth value of only the central one-third region of the bounding box as the distance value. In addition, real-time data is provided in the ROS package, using rospy.Time.Now(). Therefore, the real-time speed can be calculated based on the distance difference and the elapsed time. The depth camera is accurate down to the centimeter level. In our study, the speed is measured in meters per second. The results are displayed in **Figure 8**.

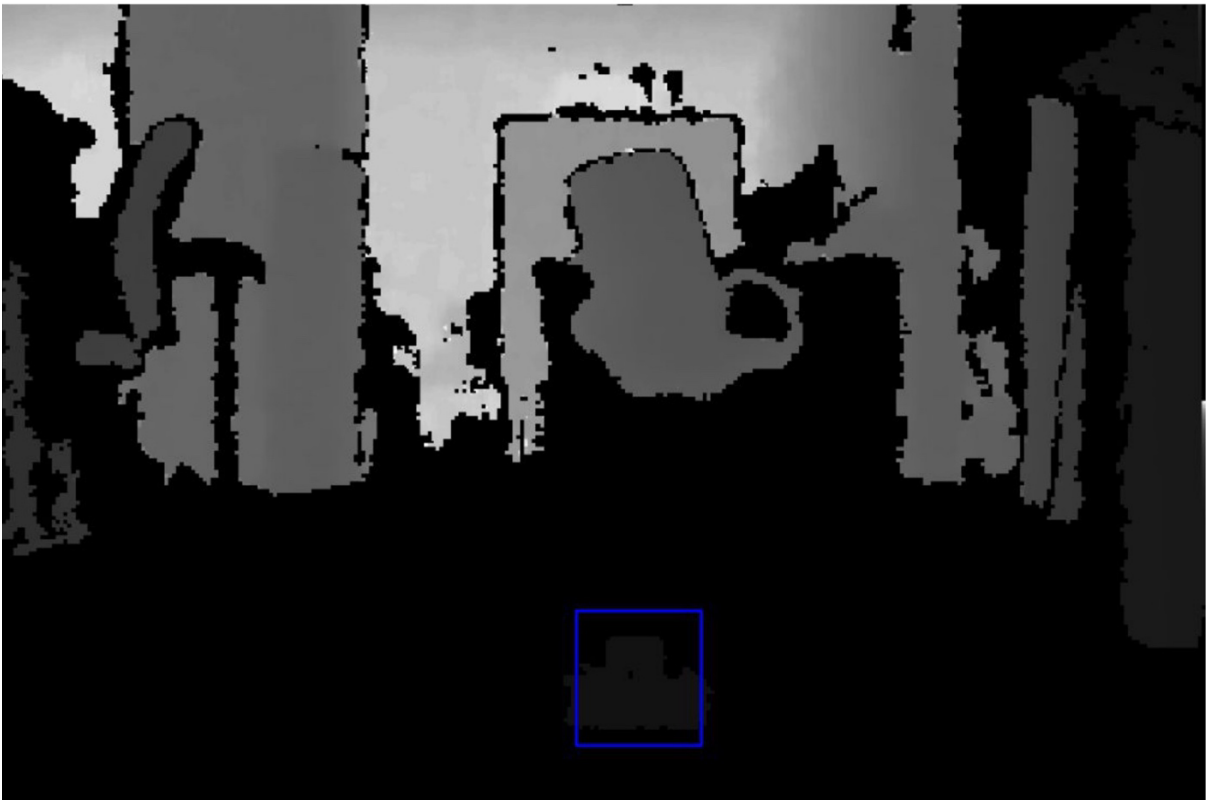

**Figure 7 Project Bounding Box in Depth Image to Calculate Distance**

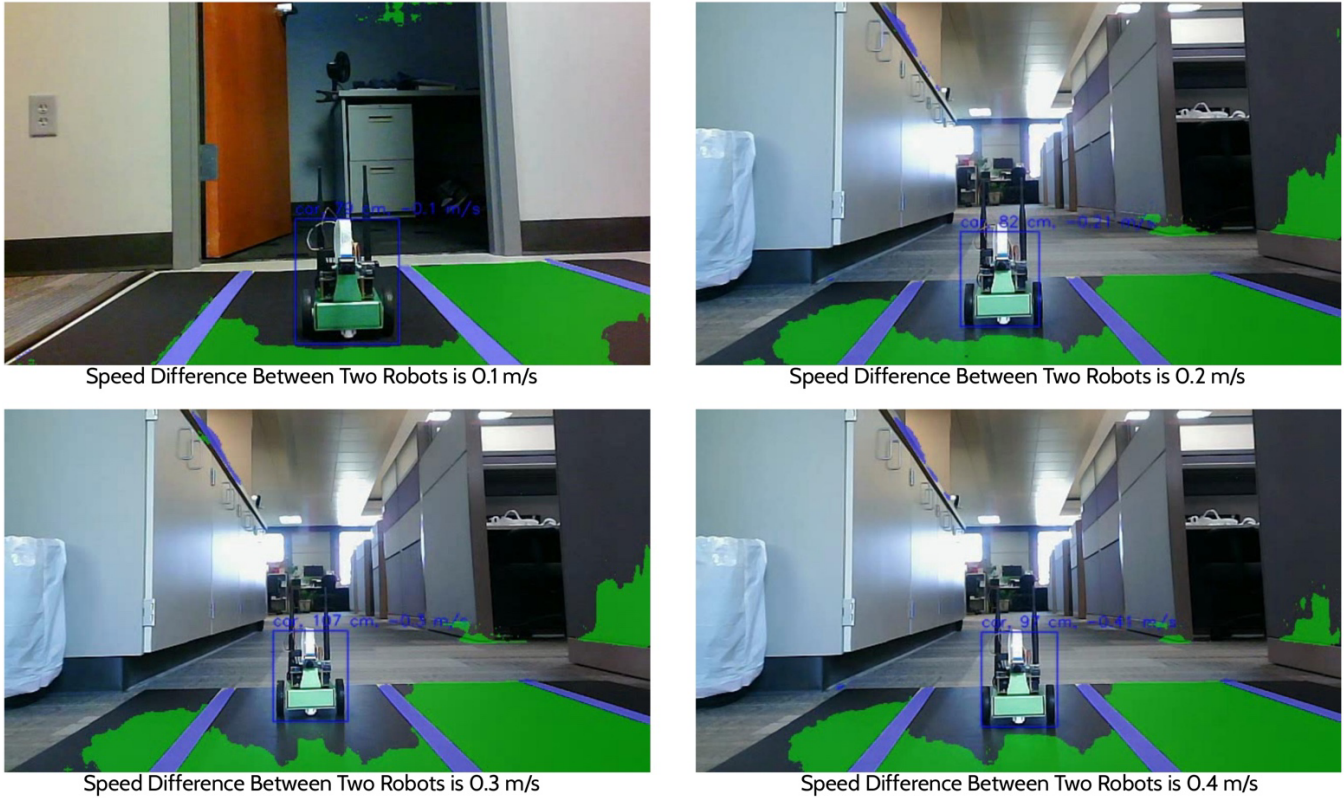

**Figure 8 Oncoming Vehicle Detection and Acquisition of Real-Time Distance and Speed**

## RESULT AND DISCUSSION

In the test, we set the speed difference between the two robots to range from 0.1 meters per second to 0.4 meters per second. For each speed difference, we conducted four runs and saved the live stream displayed by the camera as a video. This allows us to evaluate the accuracy of the speed measurements later. Then, the parameters average speed difference and mean squared error (MSE) are used to evaluate the system performance. MSE is a metric that measures the accuracy of a predictive model. From **Table 2**, with the set speed difference of 0.1 meters per second, 0.2 meters per second, and 0.3 meters per second, the measured mean speed differences are 0.105 meters per second, 0.235 meters per second, and 0.358 meters per second, respectively, indicating a close approximation to the set values. Additionally, the MSE values are also very small. However, when the speed difference reaches 0.4 meter per second, the calculated average speed shows a significant bias. The main reason is due to the relatively poor accuracy of the depth camera.

**Table 2 Evaluation of the Real-Time Speed**

| Speed Difference (In Setting) | Mean (Real World) | Mean Squared Error |
|---|---|---|
| -0.1 m/s | -0.105 | 0.0031 |
| -0.2 m/s | -0.235 | 0.0077 |
| -0.3 m/s | -0.358 | 0.0079 |
| -0.4 m/s | -0.566 | 0.0497 |

In future TMA field tests, the depth camera should be replaced with high-accuracy distance measurement sensors, such as LiDAR, to measure real-time distance and speed more accurately. The color camera and LiDAR need to be calibrated. The node integrated with the fine-tuned YOLOP algorithm will continue to subscribe to the color image topic and LiDAR topic to perform the vehicle localization, obtain the real-time distance, and further calculate real-time speed. If real-time distance and speed are available, the system will calculate the stopping sight distance based on the real-time speed. If the calculated stopping distance is less than the recommended value in the AASHTO standard (as shown in **Table 1**), the system will activate additional flashing warning LEDs to alert the driver. These flashing LEDs will prompt the inattention driver to perform an emergency brake and switch lanes.

## CONCLUSION REMARKING

In this study, we aim to development a fully automated proactive warning system and apply it to TMA/ATMA to enhance the safety of mobile work zones. Before deploying the entire system for field testing, we need to demonstrate its feasibility. Therefore, a laboratory environment simulation has been set up. In the simulation, two Jetson Nano robots and desktop GPU are used. One Jetson Nano simulates an oncoming vehicle, while the other, equipped with a color camera and depth camera, simulates the TMA follower truck. Leveraging ROS's distributed computing feature, the desktop GPU serves as the computation center. New datasets prepared in the laboratory environment are used to train the validate the YOLOP algorithm. Computer vision techniques, such as median blur and connected components, are implemented to determine the lane of the oncoming vehicle. The depth camera provides real-time distance which can be used to calculate real-time speed. In the simulation, if the distance is less than 0.3 meter, flashing warning LEDs are activated. Testing with various speed difference showed small MSE values for speed differences between 0.1 and 0.3 meters per second. However, at a speed difference of 0.4 meter per second, the calculated average speed showed significant bias, mainly due to the relatively poor accuracy of the depth camera. For future field tests, the depth camera should be replaced with LiDAR to obtain accurate depth information. By incorporating the AASHTO Green Book's stop sight distance (SSD) standards, if the real-time distance is less than the SSD for the current speed, additional flashing LED warnings will alert drivers to change lanes or conduct emergency brake. ROS's distributed computing capabilities also enhance system deployment flexibility and cost efficiency.

## AUTHOR CONTRIBUTIONS

The authors confirm contribution to the paper as follows: study conception and design: Yu, Adu-Gyamfi; data collection: Yu; analysis and interpretation of results: Yu, Zhang, Adu-Gyamfi; draft manuscript preparation: Yu, Adu-Gyamfi. All authors reviewed the results and approved the final version of the manuscript.